\documentclass[preprint,12pt,authoryear]{elsarticle}
\usepackage{amssymb}
\usepackage{amsmath}
\usepackage{graphicx}
\usepackage{amsmath, amssymb}
\usepackage{hyperref}
\usepackage{float}
\usepackage[utf8]{inputenc}
\usepackage[T5]{fontenc}
\usepackage{xcolor}
\usepackage{caption}
\usepackage[justification=centering]{caption}
\usepackage{tabularx}
\usepackage{array}
\usepackage{booktabs} 
\usepackage{lscape}
\usepackage{orcidlink}

\usepackage{longtable}   
\usepackage{adjustbox}
\usepackage{multirow}
\usepackage{subcaption}

\begin{document}

\begin{frontmatter}

\title{Deep Feature Optimization for Enhanced Fish Freshness Assessment}

\author{Phi-Hung Hoang}
\ead{hunghpde180523@fpt.edu.vn}

\author{Nam-Thuan Trinh}
\ead{thuantnde180305@fpt.edu.vn}

\author{Van-Manh Tran}
\ead{manhtvde180090@fpt.edu.vn}

\author{Thi-Thu-Hong Phan\corref{cor}\orcidlink{0000-0001-6880-3721}}
\ead{hongptt11@fe.edu.vn}

\cortext[cor]{Corresponding author}

\affiliation{organization={Department of Artificial Intelligence, FPT University},
            addressline={Da Nang},
            postcode={550000}, 
            country={Vietnam}}

\begin{abstract}
Assessing fish freshness is vital for ensuring food safety and minimizing economic losses in the seafood industry. However, traditional sensory evaluation remains subjective, time-consuming, and inconsistent. Although recent advances in deep learning have automated visual freshness prediction, challenges related to accuracy and feature transparency persist. This study introduces a unified three-stage framework that refines and leverages deep visual representations for reliable fish freshness assessment. First, five state-of-the-art vision architectures--ResNet-50, DenseNet-121, EfficientNet-B0, ConvNeXt-Base, and Swin-Tiny--are fine-tuned to establish a strong baseline. Next, multi-level deep features extracted from these backbones are used to train seven classical machine learning classifiers, integrating deep and traditional decision mechanisms. Finally, feature selection methods based on Light Gradient Boosting Machine (LGBM), Random Forest, and Lasso identify a compact and informative subset of features. Experiments on the Freshness of the Fish Eyes (FFE) dataset demonstrate that the best configuration--combining Swin-Tiny features, an Extra Trees classifier, and LGBM-based feature selection--achieves an accuracy of 85.99\%, outperforming recent studies on the same dataset by 8.69--22.78\%. These findings confirm the effectiveness and generalizability of the proposed framework for visual quality evaluation tasks.
\end{abstract}


\end{frontmatter}

\section{Introduction}

Fish is an essential component of the global diet, providing high-quality protein and vital nutrients that contribute significantly to human health~\citep{Tidwell2001,Chen2022}. Beyond its nutritional value, the fisheries sector also plays a crucial economic role in global food security and trade. However, fish is an inherently perishable commodity, and improper handling or storage can rapidly deteriorate its freshness, leading to both economic losses and potential health risks. Traditionally, freshness evaluation has relied on sensory inspection by trained experts, who assess the eyes, gills, skin, and odor. Although practical, these subjective methods are time-consuming, inconsistent, and difficult to scale for large quantities of fish, particularly under industrial conditions~\citep{Hassoun2017, Prabhakar2020}. Therefore, developing rapid, accurate, and automated techniques for assessing fish freshness has become essential to ensure food safety and improve management efficiency in the seafood industry.

To address these challenges, researchers have explored a wide range of image-based approaches for objective and automated evaluation. Early studies primarily employed handcrafted visual descriptors--such as color features, local binary patterns (LBP), and gray-level co-occurrence matrices (GLCM)--combined with machine learning classifiers to infer freshness levels~\citep{Medeiros2021, Syarwani2022, Arora2022, Hoang2025}. While these methods achieved promising results, their performance heavily depended on manual feature design and often lacked robustness across diverse imaging conditions.

The advent of deep learning has revolutionized visual analysis by enabling automatic feature learning directly from raw images. Several convolutional neural network (CNN)-based architectures and lightweight variants have been applied to fish freshness prediction~\citep{Prasetyo2022, Jayasundara2023, Cahyo2024, Khaleel2024, Sanga2024}. However, despite clear progress, the Freshness of the Fish Eyes (FFE) dataset remains particularly challenging, as existing deep models have achieved only moderate accuracies of 63\% and 77\% in the studies of~\cite{Prasetyo2022, Yildiz2024}. This suggests that distinguishing subtle inter-class differences in eye appearance still demands more discriminative and interpretable representations.

Moreover, existing studies often evaluate deep networks in an end-to-end manner, without explicitly analyzing how feature abstraction levels or feature selection mechanisms affect performance.
For instance,~\cite{Prasetyo2022}  directly used CNN or pre-trained models for classification without optimizing or interpreting the extracted embeddings, while~\cite{Yildiz2024} focused solely on feature extraction from CNNs without systematic dimensionality refinement.

In this context, this study introduces a comprehensive deep feature optimization framework that systematically evaluates and refines learned representations for fish freshness assessment. Specifically, we:

\begin{itemize}

\item  Evaluate several state-of-the-art vision backbones--including ResNet-50, DenseNet-121, EfficientNet-B0, Swin-Tiny, and ConvNeXt-Base--to establish a robust baseline for image-based freshness classification.

\item  Apply Grad-CAM visualization to interpret the visual cues guiding model decisions.

\item Extract deep embeddings from different abstraction levels of each backbone and evaluate their discriminative power using classical machine learning classifiers.

\item  Investigate embedded feature selection methods--LGBM (boosting), Random Forest (bagging), and Lasso (L1 regularization)--to identify compact and discriminative subsets of features.

\end{itemize}

To the best of our knowledge, this is the first study to systematically analyze how deep feature abstraction and embedded feature optimization jointly affect fish freshness classification performance on the public FFE dataset. The proposed framework demonstrates improved accuracy and enhanced interpretability, while maintaining computational efficiency suitable for practical applications.

The rest of the paper is organized as follows. Section~\ref{sec:related_works} reviews related literature. Section~\ref{sec:method} presents the proposed methodology, covering the main approaches and techniques used. Section~\ref{sec:experiments} describes the experimental setup. Section~\ref{sec:results_and_discussion} presents the results and offers a comprehensive discussion of the findings. Finally, Section~\ref{sec:conclusion} concludes the study and suggests directions for future research.

\section{Related works}
\label{sec:related_works}

Early computer vision approaches for fish freshness assessment primarily relied on manually engineered features extracted from digital images to capture visual degradation in the eyes, gills, and skin. Color-based descriptors were dominant, often computed as statistical measures or histograms in RGB, HSV, HSI, and CIELAB spaces to quantify discoloration.
For instance,~\cite{Medeiros2021} extracted a comprehensive set of colorimetric parameters from multiple color spaces and achieved 100\% accuracy in classifying tuna and salmon freshness using an AutoML framework. Texture features like the GLCM and LBP were also used to represent surface variations such as eye cloudiness.~\cite{Syarwani2022} combined HSV color features with GLCM descriptors, attaining 94.28\% accuracy for Nile Tilapia freshness using a SVM.~\cite{Arora2022} further advanced feature fusion by computing a unified ``Q-score'' from weighted visual features of the gills, eyes, and skin, achieving 98.07\% accuracy. More recently,~\cite{Hoang2025} demonstrated that a systematic fusion of handcrafted color and texture features could achieve 77.56\% accuracy on the FFE dataset, highlighting the potential of optimized feature integration for improving machine learning-based fish freshness classification.

Building on these foundations, deep learning approaches were developed to extract discriminative features automatically. Lightweight architectures such as MobileNetV1 bottleneck with Expansion (MB-BE) achieved 63.21\% accuracy on the FFE dataset, illustrating the trade-off between efficiency and performance~\citep{Prasetyo2022}.~\cite{Jayasundara2023} developed FishNET-S for Indian Sardinella focusing on fish eyes and FishNET-T for Yellowfin Tuna focusing on fish meat, achieving 84.1\% and 68.3\% accuracy, respectively. Concurrently, many studies focused on well-established CNN architectures such as ResNet and DenseNet. By leveraging pre-trained models and data fusion techniques, these approaches reported exceptional classification accuracies ranging from approximately 93\% to 100\%~\citep{Cahyo2024, Khaleel2024, Sanga2024}.

Recent research has explored advanced hybrid models that combine multiple neural network architectures to improve fish freshness assessment.~\cite{Rodrigues2024} were the first to apply Vision transformers (ViT) in this context. Their two-stage system first segmented the fish eye region with high performance, achieving a 98.77\% detection rate and 85.7\% IoU using a Segformer, before classification with a ViT resulted in 80.8\% accuracy. Hybrid architectures have also been explored, with~\cite{Biswas2025} proposing a CNN-LSTM model that integrates spatial and sequential information and uses LIME for interpretability, achieving 86\% accuracy.~\cite{Peries2025} conducted a systematic comparison of multiple pre-trained architectures across whole fish, fish eyes, and fish gills, with the highest accuracy of 99.13\% achieved on gills using DenseNet121.

Hybrid frameworks have become an important direction in automated fish freshness assessment, employing deep learning to generate rich feature representations and traditional machine learning to perform classification. This combination leverages the strengths of both approaches, balancing accuracy with efficiency and interpretability. For example,~\cite{Kılıcarslan2024} extracted features from pre-trained models and classified them with traditional algorithms, achieving a 100\% success rate. Similarly,~\cite{Yildiz2024} applied pre-trained CNNs to the FFE dataset and found that combining VGG19 features with an ANN yielded the highest accuracy of 77.3\%. Extending this concept,~\cite{Lanjewar2024} employed a NasNet-LSTM architecture for feature extraction along with data balancing and feature selection techniques, achieving Matthew's correlation coefficient (MCC) and Cohen's kappa coefficient (KC) scores of 99.1\%.

Another advanced strategy is multimodal data fusion, which integrates information from different sensors to provide a more comprehensive representation of freshness.~\cite{Hardy2024} fused fluorescence, visible, and near-infrared spectroscopy data, reporting nearly perfect accuracies of 99.5\% compared to single-mode analyses of 77.1\%. Similarly,~\cite{Balım2025} combined RGB images with laser reflectance data, achieving 88.44\% accuracy.

A review of the literature also highlights several common limitations, as summarized in Table ~\ref{tab:limitations}. Many studies rely on small or private datasets collected under ideal laboratory conditions, such as controlled lighting and background, which hinders fair comparison and raises concerns about generalizability. Methodologically, much of the research focuses on a single anatomical region such as the eye or relies exclusively on either handcrafted features or deep learning, rarely exploring a systematic combination of both. Furthermore, some hybrid frameworks use conventional CNN architectures that may not fully capture complex visual cues.

{\footnotesize
\begin{longtable}{|p{0.14\textwidth}|p{0.25\textwidth}|p{0.18\textwidth}|p{0.32\textwidth}|}
\caption{Summary of recent studies on fish freshness assessment} \label{tab:limitations} \\
\hline
\textbf{Studies} & \textbf{Methodology} & \textbf{Accuracy} & \textbf{Limitations} \\ \hline
\endfirsthead
\multicolumn{4}{c}
{{\tablename\ \thetable{} -- continued from previous page}} \\
\hline
\textbf{Studies} & \textbf{Methodology} & \textbf{Accuracy} & \textbf{Limitations} \\ \hline
\endhead
\hline
\endlastfoot

~\cite{Medeiros2021} & Colorimetric features from multiple spaces classified with AutoML & 100\% & Limited number of tuna and salmon samples. Freshness was assessed only using color features, which may overlook other important indicators\\ \hline

~\cite{Syarwani2022} & HSV + GLCM features with SVM classifier & 94.28\% & Small sample of Nile Tilapia eyes and gills in lab conditions\\ \hline

~\cite{Arora2022} & Fused gill, eye, and skin features into a single Q-score & 98.07\% & Limited Rohu data reduces generalizability across species and environments. Segmentation employed simple traditional image processing\\ \hline


~\cite{Prasetyo2022} & Proposed a lightweight CNN (MB-BE) & 63.21\% & Low overall accuracy due to lightweight design and limited feature representation and generalization\\ \hline

~\cite{Jayasundara2023} & Proposed two specialized CNNs: FishNET-S (for small fish eyes) and FishNET-T (for large fish meat) & 90\% (FishNET-S), 100\% (FishNET-T) & Two separate models for different fish species limit generalization and controlled data reduce robustness in real-world settings\\ \hline

~\cite{Cahyo2024} & ResNet101 + GLCM features & 100\% & The research focused on a single species with limited data, employing conventional image processing for segmentation in controlled lab lighting\\ \hline

~\cite{Khaleel2024,Sanga2024} & ResNet18 + image fusion; Compared VGG19, MobileNetV2, DenseNet201, ResNet50 & 93\%, 100\% & Considering only two freshness classes, the deep learning model's interpretability remains limited\\ \hline

~\cite{Rodrigues2024} & Segformer for segmentation + ViT for classification & 98.77\% detection, 85.7\% IoU, 80.8\% accuracy & Using transformers for segmentation and classification requires high computational resources\\ \hline

~\cite{Biswas2025} & CNN-LSTM integrating spatial–sequential features with LIME interpretability & 86\% & The integration of CNN and LSTM on a single-species dataset with binary classification increases model complexity\\ \hline

~\cite{Peries2025} & Evaluated CNNs and pre-trained models on fish eyes, gills, and whole fish. & 99.13\% (DenseNet121, gill) & Low image resolution, binary classification, single-region data, and limited explainability reduce generalization\\ \hline

~\cite{Kılıcarslan2024} & Features from MobileNetV2, Xception, VGG16 + ML classifiers (SVM, LR, ANN, RF) & 100\% & Single-species dataset under controlled conditions, binary classification, and limited explainability\\ \hline

~\cite{Yildiz2024} & VGG19 \& SqueezeNet features + ML classifiers (KNN, RF, SVM, LR, ANN) & 77.3\% & Extracted features are insufficient to capture complex, long-range dependencies in the data\\ \hline

~\cite{Lanjewar2024} & NasNet–LSTM hybrid with SMOTEENN balancing and feature selection & 99.1\% (MCC \& KC scores) & Focusing only on the eye region and using a CNN–LSTM hybrid, the model becomes more complex and harder to interpret\\ \hline

~\cite{Hardy2024} & Fused fluorescence + VIS–NIR data via LDA, KNN, and ensemble bagged trees & 99.5\% & Limited fish samples, generous classification metric, controlled lab environment, and single-species focus\\ \hline

~\cite{Balım2025} & Laser + CNN feature fusion with SVM, MLP, RF classifiers & 88.4\% & The evaluation used only mackerel and a 940 nm laser wavelength, making generalization to other species and conditions difficult\\ \hline
\end{longtable}
}

These limitations highlight the need for a systematic framework that not only evaluates deep feature representations across different abstraction levels but also incorporates feature optimization techniques to achieve compact, interpretable, and generalizable freshness classification.

\section{Methodology}
\label{sec:method}

\subsection{Overview of the proposed approach}

\begin{figure}[H]
   \centering
   \hspace*{-1.4cm}
   \includegraphics[width=1.2\linewidth]{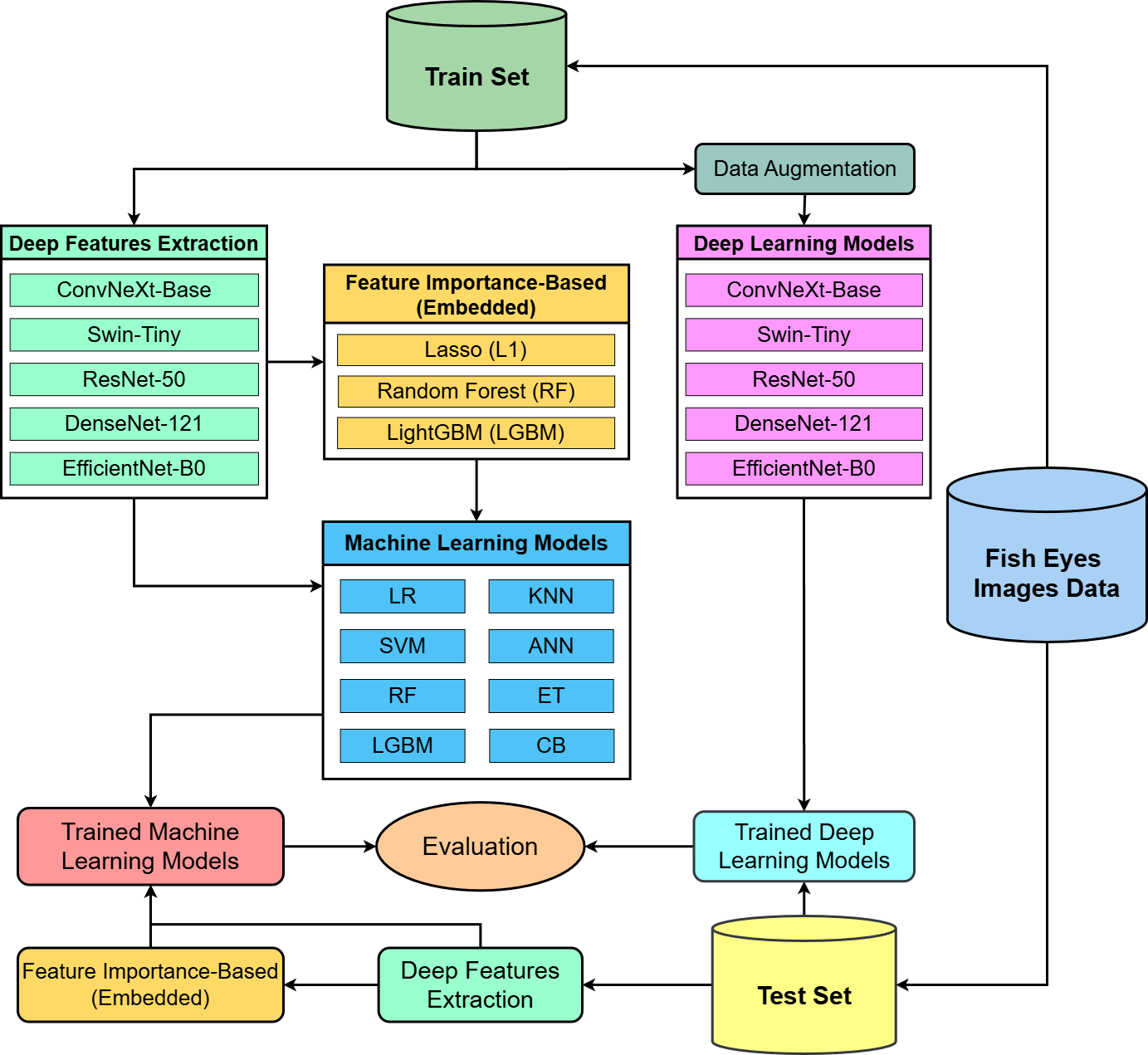}
   \caption{Overview of the proposed methodology}
   \label{fig:pipeline}
\end{figure}

To systematically evaluate and overcome the inherent difficulty of accurately classifying fish freshness from imaging data, this study proposes a novel and progressive framework that integrates modern deep learning and machine learning techniques within a unified training–evaluation pipeline (Figure~\ref{fig:pipeline}). The framework is designed not merely as a comparative setup but as a structured exploration of how recent advances in vision architectures and hybrid learning strategies can be effectively exploited for fine-grained classification tasks.

In the first stage, a series of representative deep learning models--ResNet-50, DenseNet-121, EfficientNet-B0, Swin-Tiny, and ConvNeXt-Base--were fine-tuned using Random search for hyperparameter optimization. These architectures were deliberately chosen to capture diverse design philosophies, ranging from conventional convolutional networks to transformer-based and convolution–transformer hybrid models. This phase aimed to establish a robust baseline and to investigate how architectural diversity and the Random search strategy influence the representation of visual freshness cues.

In the second stage, the study transitioned to a hybrid deep feature–based learning framework. Rather than relying solely on fully abstract representations, deep features were extracted from the intermediate-to-high abstraction levels of each fine-tuned backbone, specifically from the global average pooling (GAP) outputs.
Instead of combining these representations, the discriminative power of features from each stage was individually evaluated to determine which level of abstraction yields superior classification performance.
This design choice was motivated by the characteristics of fish-eye images, where freshness cues depend on both fine-grained visual details (e.g., texture and glossiness) and higher-level semantic attributes (e.g., overall color consistency and shape). Evaluating both levels helped identify the most informative representation that balances local texture information with global semantic cues, making it particularly suitable for this application. 
Two complementary analysis pathways were then explored:

(1) Full deep features, where all extracted representations were directly used for classification; and 

(2) Feature-selected deep features, where redundant or less informative dimensions were removed using embedded feature selection methods--LASSO (L1 regularization), RF (bagging), and LGBM (boosting)--to automatically rank and retain the most discriminative subset of features.

This design enabled a systematic comparison between the raw representational capacity and the compact, feature-selected variants, highlighting the trade-offs between richness and efficiency of learned embeddings.

The resulting feature sets (either full or feature-selected) were then used as inputs to a suite of classical machine learning algorithms, including Logistic Regression (LR), K-Nearest Neighbors (KNN), Support Vector Machine (SVM), Random Forest (RF), Extra Trees (ET), LightGBM (LGBM), and CatBoost (CB). This stage examined whether explicit decision mechanisms could further enhance discriminative power beyond the softmax-based classification of deep networks.

In the final evaluation stage, both fine-tuned deep models and ML classifiers trained on deep features were assessed under identical conditions using standard performance metrics. Grad-CAM visualization was employed to interpret spatial attention in deep models, offering a comprehensive view of model behavior and interpretability.

The following section presents a detailed description of the methods used in this study.

\subsection{Deep learning models}

To evaluate the performance and efficiency of various deep learning architectures, our study includes a selection of prominent models representing significant advancements in visual recognition. Training deep neural networks effectively presents challenges, particularly the degradation of performance with increasing depth due to optimization difficulties like vanishing gradients. The architectures selected have each introduced novel concepts or refinements to overcome these hurdles and achieve state-of-the-art results.

\subsubsection{ResNet-50}

ResNet~\citep{ResNet50} introduces residual learning to address the optimization challenges of very deep neural networks. Instead of learning direct mappings, it employs shortcut connections that enable residual functions, improving gradient flow and convergence. The ResNet-50 architecture consists of 50 layers arranged in four residual stages with downsampling at the start of stages 2, 3, and 4. Each bottleneck block includes $1\times1$, $3\times3$, and $1\times1$ convolutions, while the network begins with a $7\times7$ convolution and max pooling, and concludes with global average pooling and a fully connected classification layer.

\subsubsection{DenseNet-121}

Dense Convolutional Networks (DenseNets)~\citep{DenseNet2017} introduce dense connectivity to enhance feature reuse and parameter efficiency. Instead of summing residuals as in ResNet, each layer in a DenseNet concatenates its output with all preceding feature maps within the same block. DenseNet-121 applies this principle across multiple dense blocks separated by transition layers that downsample and reduce channels using $1\times1$ convolutions and average pooling. Each dense layer consists of batch normalization, ReLU activation, and convolution, followed by global average pooling and a fully connected classification layer.

\subsubsection{EfficientNet-B0}

EfficientNet~\citep{EfficientNet2019} introduces a compound scaling method that uniformly balances network depth, width, and input resolution using learned coefficients, addressing the inefficiency of scaling a single dimension. EfficientNet-B0 serves as the baseline architecture discovered through neural architecture search, optimized for both accuracy and computational cost. Its primary building block, the MBConv, employs an inverted bottleneck with channel expansion, depthwise separable convolution, and Squeeze-and-Excitation attention. The model follows a standard multi-stage ConvNet structure, concluding with global average pooling and a fully connected classification layer.

\subsubsection{Swin Transformer-Tiny (Swin-T)}

The Swin Transformer~\citep{Swin2021} introduces a hierarchical Vision Transformer that improves efficiency for high-resolution and dense prediction tasks. It builds hierarchical feature maps through patch merging and applies window-based self-attention within local regions for linear computational complexity. The Swin Transformer-Tiny (Swin-T) variant uses shifted windows to capture cross-window dependencies, with patch merging between stages to reduce tokens and expand feature dimensions before global pooling and classification.

\subsubsection{ConvNeXt-Base}

ConvNeXt~\citep{ConVneXt2022} revisits conventional ConvNet architectures by incorporating design principles from Vision Transformers to achieve comparable performance while maintaining convolutional efficiency and inductive biases. ConvNeXt-Base follows a ResNet-like stage structure with key modifications, including a larger-stride stem, non-overlapping patch-like convolutions, large-kernel depthwise layers, and inverted bottlenecks inspired by Transformer MLPs. It replaces batch normalization with layer normalization and applies downsampling between stages, forming a hierarchical yet purely convolutional architecture.

\subsection{Deep feature extraction strategy}
\label{sec:deep_feature_extraction_strategy}

The effectiveness of deep features hinges on a crucial trade-off: mid-level layers retain local, fine-grained spatial details necessary for subtle cues, such as texture and glossiness, while high-level layers capture richer, more abstract semantic patterns. To systematically identify the most informative representations for assessing fish freshness, features from both mid-level and high-level stages of each fine-tuned model were independently extracted. These representations were subsequently condensed using GAP, a standard technique that creates compact, fixed-length vectors from feature maps. This approach preserves essential global information while significantly reducing feature dimensionality and computational complexity for the downstream traditional machine learning models. The specific candidate extraction points for each model, along with their detailed rationale, are summarized in Table~\ref {tab:feature_extraction_strategy}.  

\renewcommand{\arraystretch}{1.2}
\begin{table}[H]
\centering
\footnotesize
\caption{Candidate deep feature extraction points for comparative analysis.}
\label{tab:feature_extraction_strategy}
\begin{tabularx}{\textwidth}{|p{0.15\textwidth}|p{0.15\textwidth}|X|p{0.15\textwidth}|X|}
\hline
\textbf{Model} & \textbf{Mid-level Extraction} & \textbf{Rationale} & \textbf{High-level Extraction} & \textbf{Rationale} \\ \hline
ResNet-50 & Layer 3 (1024-dim) & Captures mid-level patterns before final downsampling & Layer 4 (2048-dim) & Encodes abstract semantic features \\ \hline
DenseNet-121 & DenseBlock 3 (1024-dim) & Rich fused representation from interconnected layers & DenseBlock 4 (1024-dim) & Final refined feature set before output \\ \hline
EfficientNet-B0 & Block 6 (192-dim) & Captures complex spatial patterns with larger receptive field & Block 7 (320-dim) & Refines features for high-level representation \\ \hline
Swin-Tiny & Stage 3 (384-dim) & Contextual features from self-attention layers & Stage 4 (768-dim) & Captures global representation of image tokens \\ \hline
ConvNeXt-Base & Stage 3 (512-dim) & Aggregated features from deep ConvNeXt blocks & Stage 4 (1024-dim) & Highest-level feature hierarchy \\ \hline
\end{tabularx}
\end{table}

\subsection{Explainability with Grad-CAM}

Grad-CAM (Gradient-weighted Class Activation Mapping) is a visualization technique used to interpret the decision-making process of deep learning models by highlighting the important regions in the input image that contribute most to the model's prediction~\citep{Gradcam2017}. Grad-CAM computes the gradients of the target class score with respect to the feature maps of the last convolutional layer (in CNNs) or the corresponding representation layer (in Transformers) to generate a heatmap indicating the areas of focus.

In this study, Grad-CAM is applied to both CNN and Transformer-based models to better understand which image regions influence their classification decisions. For CNNs, the heatmaps reveal spatial regions of interest based on convolutional feature maps, while for Transformers, Grad-CAM is adapted to visualize attention in relevant layers or tokens. This helps provide interpretability and validates that the models are focusing on meaningful visual cues related to fish freshness assessment.

\subsection{Embedded feature selection strategy}

\begin{figure}[H]
    \centering
    \includegraphics[width=0.8\linewidth]{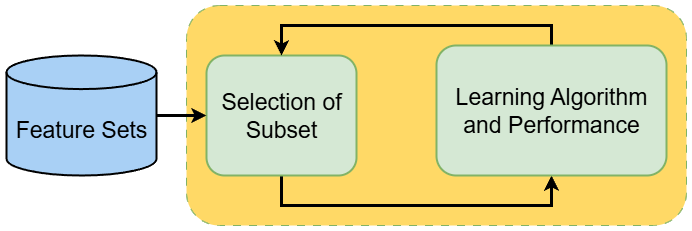}
    \caption{Mechanism of embedded feature selection in model training.}
    \label{fig:embedded_method}
\end{figure}

In this study, we adopt embedded feature selection as our primary approach. Embedded methods integrate the assessment of feature relevance directly into the model training process, unlike filter and wrapper techniques, which treat feature selection as a separate preprocessing step. By evaluating predictive value during model construction (Figure \ref{fig:embedded_method}), embedded methods produce a compact, model-specific subset of features that enhances both robustness and generalization.
Our choice of embedded feature selection is motivated by prior evidence demonstrating its effectiveness across diverse domains. For example, in water quality index (WQI) prediction \citep{Bui2023_WQI} and rice seed purity classification \citep{Phan2025_RiceSeedPurity}, embedded techniques consistently outperformed filter and wrapper approaches in terms of predictive performance and feature selection capability. Considering the high dimensionality and complexity of deep features extracted for fish freshness assessment, embedded methods are expected to provide a reliable and efficient strategy for selecting the most informative features.

To perform comprehensive feature optimization, we employ three complementary embedded methods: Lasso regression (L1) - Regularization), RF, and LGBM. These techniques cover the major mechanisms of embedded feature selection:

\vspace{0.2cm}

\textbf{\textit{{1. Tree-based ensemble selection (Bagging \& Boosting)}}}
\begin{itemize}
    \item \textbf{Random forest (Bagging):} RF averages importance across many independent decision trees. Features are ranked based on their ability to improve node purity (e.g., Gini impurity), providing a stable and generalized measure of relevance.
    \item \textbf{LightGBM (Boosting):} LGBM sequentially builds trees, focusing on features that provide the greatest Gain (reduction in loss) at each split. This effectively captures complex non-linear relationships and the predictive power of features.
\end{itemize}

\textbf{\textit{{2. Regularization-based selection (Lasso regression (L1):}}}
Lasso introduces a penalty proportional to the absolute magnitude of coefficients. The L1 penalty drives coefficients of less informative features to exactly zero, effectively performing selection. This is particularly effective for managing multicollinearity in high-dimensional feature spaces.

By employing three embedded feature selection methods-Lasso, RF, and LGBM-our approach ensures that the selected feature subsets are optimized, stable, and minimally redundant for fish freshness classification. For each method, the selection threshold was empirically determined by testing multiple cutoffs based on feature importance scores or coefficient magnitudes.

\subsection{Machine learning models}

This study investigates the classification of fish into three freshness categories (``\textit{Highly Fresh}'', ``\textit{Fresh}'', ``\textit{Not Fresh}'') using a diverse set of machine learning algorithms. The models were selected based on their ability to handle structured data, computational efficiency, and capability to capture both linear and complex non-linear relationships. A detailed description of each model is presented below.

\textbf{Logistic Regression (LR):} is a statistical model that uses a logistic function to predict the probability of a binary outcome \citep{Hosmer2000}. It transforms a linear combination of input features into a value between 0 and 1. The model is trained by finding the coefficients that maximize the likelihood of the observed data, and these coefficients indicate the influence of each feature on the predicted probability.

\textbf{K-Nearest Neighbors (KNN):} is a simple yet effective non-parametric algorithm for classification and regression~\citep{Fix1951}. It classifies a sample by identifying its $k$ closest neighbors based on a distance metric such as Euclidean or Manhattan distance and assigning the majority label. By relying on local data structure, KNN captures patterns without assuming any specific underlying distribution.

\textbf{Support Vector Machines (SVM):} is a supervised learning algorithm that identifies the optimal hyperplane separating classes with maximum margin~\citep{Vapnik1995}. The closest data points, known as support vectors, define this boundary. For non-linear relationships, kernel functions project data into higher-dimensional spaces, enabling effective classification of complex patterns.

\textbf{Artificial Neural Networks (ANN):} are computational models inspired by biological neural systems~\citep{McCulloch1943}. They consist of interconnected layers of neurons that transform inputs through weighted connections and non-linear activations. Trained via backpropagation to minimize prediction error, ANNs can model complex non-linear relationships, enabling effective performance across classification and regression tasks.

\textbf{Random Forest (RF):} is an ensemble learning algorithm that builds a collection of decision trees to improve classification performance~\citep{Breiman2001}. Each tree is trained on a random subset of the data and a random subset of features, which introduces diversity and reduces overfitting. To classify a new sample, each tree makes a prediction, and the final class is determined by majority voting among all trees. This approach allows Random Forest to capture complex patterns and interactions in the data while maintaining high accuracy and robustness.

\textbf{Extra Trees (ET):} is an ensemble learning algorithm similar to Random Forest that constructs a collection of decision trees~\citep{Geurts2006}. Unlike Random Forest, Extra Trees select split points completely at random for each feature when building the trees, which increases variability and reduces overfitting. Each tree provides a class prediction, and the final output is determined by majority voting among all trees. This method allows Extra Trees to capture complex patterns efficiently while being fast and robust.

\textbf{LightGBM (LGBM):} is a gradient boosting framework that constructs decision trees sequentially to minimize residual errors~\citep{Ke2017}. It incorporates Gradient-based One-Side Sampling (GOSS) to emphasize informative samples and Exclusive Feature Bundling (EFB) to reduce feature dimensionality. Designed for efficiency and scalability, LGBM delivers fast and accurate predictions on large datasets with mixed feature types.

\textbf{CatBoost (CB):} is a gradient boosting algorithm that constructs an ensemble of decision trees sequentially to minimize prediction errors~\citep{Dorogush2018}. It efficiently handles categorical features through ordered boosting, which mitigates overfitting and enhances generalization. By combining the outputs of all trees, CatBoost achieves high accuracy on complex datasets with both numerical and categorical variables.

\section{Experiments}
\label{sec:experiments}

\subsection{Data description}
\begin{figure}[H]
   \centering
   \includegraphics[width=1.0\linewidth]{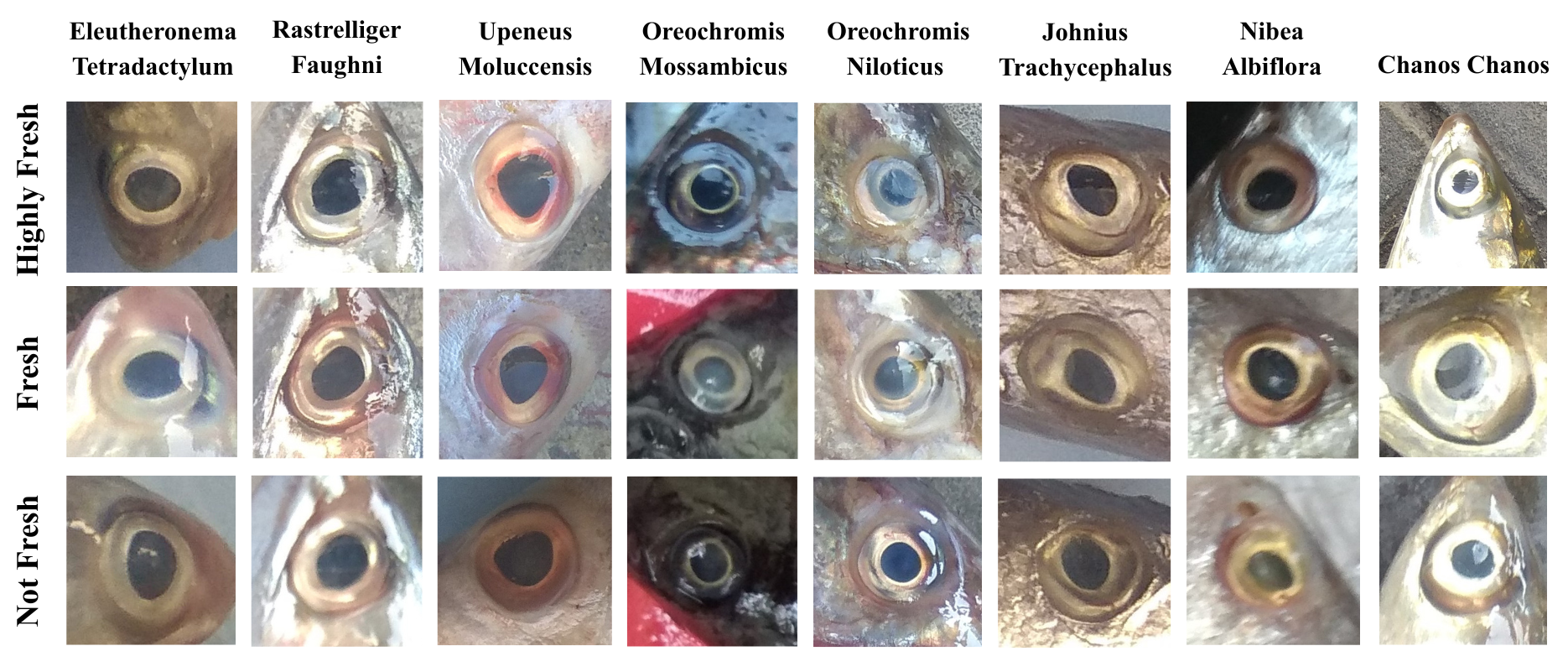}
   \caption{Sample images from the \textit{Freshness of the Fish Eyes (FFE)} dataset representing eight fish species and three freshness categories (\textit{Highly Fresh}, \textit{Fresh}, and \textit{Not Fresh}).}
   \label{fig:dataset}
\end{figure}

The dataset used in this study is the publicly available \textit{Freshness of the Fish Eyes (FFE)} collection introduced by~\cite{Prasetyo2022}. It was specifically designed for classifying fish freshness based on visual cues from the eyes. The dataset contains a total of 4,390 images categorized into three freshness levels: 1,764 ``\textit{Highly fresh}'' (days 1--2), 1,320 ``\textit{Fresh}'' (days 3--4), and 1,306 ``\textit{Not fresh}'' (days 5--6) (see Figure~\ref{fig:dataset} for sample images).  

Images were captured daily over a six-day period using a mobile phone under varied backgrounds and lighting conditions to simulate real-world scenarios. This ensures that the dataset reflects practical variations encountered in typical fish markets. The provided images have been pre-processed to focus on the eye region, which serves as the primary visual input for the models. For model development and evaluation, the dataset was split into three subsets: 64\% for training, 16\% for validation, and 20\% for testing. 

\subsection{Experimental setup}

All experiments in this study were conducted on two computational platforms. A local machine, equipped with an AMD Ryzen 7 processor, 16 GB of RAM, and an NVIDIA RTX 3050 GPU, was used for model development and classical machine learning algorithms. Computationally intensive deep learning tasks were carried out on the Kaggle platform, which provides an NVIDIA Tesla T4 GPU with 16 GB of VRAM.

The preprocessing procedures, training configurations, and optimization strategies applied in all experiments are summarized in Table~\ref{tab:experimental_settings}. This table provides a comprehensive overview of the experimental settings to ensure reproducibility and facilitate fair comparison across models.

\begin{table}[H]
\centering
\caption{Experimental settings for deep learning and classical models.}
\label{tab:experimental_settings}
\begin{tabular}{|l|p{9cm}|}
\hline
Aspect & Configuration \\
\hline
Image preprocessing & Resize: $224\times224$, Normalization: ImageNet stats \\ \hline
Data augmentation   & Horizontal flip, Rotation: ±30°, Brightness jitter \\ \hline
DL initialization   & ImageNet pre-trained weights \\ \hline
Training setup      & Optimizer: AdamW, Learning rate: 1e-4, Weight decay: 1e-2, Batch size: 64 \\ \hline
Epochs              & Max: 100, Early stopping patience: 20 \\ \hline
ML hyperparameters  & Random search optimization \\ \hline
Evaluation          & Identical conditions for all models \\ \hline
\end{tabular}
\end{table}

\section{Results and discussion}
\label{sec:results_and_discussion}

The proposed methodologies for fish freshness assessment were evaluated through five state-of-the-art deep learning models and hybrid frameworks that combine deep representations with classical classifiers. Furthermore, the influence of feature selection on model performance was analyzed. To ensure a thorough and balanced evaluation, four widely recognized metrics were employed: Accuracy (overall correctness), Precision (positive predictive value), Recall (sensitivity), and F1-Score (harmonic mean of Precision and Recall). Together, these metrics provide a comprehensive measure of both general and class-specific performance.

\subsection{Performance of deep learning models}
\label{sec:performance_of_deep_learning_models}

The experimental results of five state-of-the-art deep learning architectures revealed their strong discriminative capability in fish freshness assessment. As presented in Table \ref{tab:dl_results}, all models achieved accuracies above 80\%, confirming the robustness of deep neural networks in handling this challenging visual classification task.

Among the models, the Swin Transformer-Tiny (Swin-Tiny) achieved the highest accuracy of 84.85\%. Its transformer-based design, with the self-attention mechanism, allows the model to capture both subtle local textural details, such as the cloudiness of the fish eye lens, and broader spatial patterns in the image, which are important indicators of freshness.

\begin{table}[H]
\caption{Performance of different DL models for fish freshness assessment (\%).}
\label{tab:dl_results}
\footnotesize
\centering
\begin{tabular}{|l|c|c|c|c|}
\hline
\textbf{Model}   &    \textbf{Accuracy}   &   \textbf{Precision}   &       \textbf{Recall}     &    \textbf{F1-score}   \\
\hline
ConvNeXt-Base    &     \textbf{84.51}     &          83.73         &            83.60          &           83.62         \\
\hline
Swin-Tiny        &     \textbf{84.85}     &      \textbf{84.08}    &        \textbf{84.08}     &      \textbf{83.93}     \\
\hline
EfficientNet-B0  &         81.32          &          80.55         &            80.04          &           80.19         \\
\hline
DenseNet-121     &         80.75          &          79.91         &            80.04          &           79.97         \\
\hline
ResNet-50        &         80.07          &          79.21         &            79.59          &           79.33         \\
\hline
\end{tabular}
\end{table}

The ConvNeXt-Base model reached a closely similar accuracy of 84.51\%. This strong performance can be attributed to its modernized convolutional design, which incorporates principles inspired by vision transformers. Features such as larger kernel sizes, updated block structures, and enhanced normalization strategies allow ConvNeXt to capture both fine-grained textures and larger spatial patterns effectively, resulting in performance comparable to transformer-based models.

EfficientNet-B0, DenseNet-121, and ResNet-50 are considered mid-level models, achieving accuracies of 81.32\%, 80.75\%, and 80.07\%, respectively. EfficientNet-B0 leverages compound scaling to uniformly adjust depth, width, and resolution, capturing rich visual features efficiently. Its MBConv blocks with squeeze-and-excitation further enhance feature representation by emphasizing informative channels. DenseNet-121 and ResNet-50 rely on conventional architectures that effectively extract general features but may be limited in modeling complex spatial relationships and subtle textures, which are important for fine-grained freshness assessment.

\subsubsection{Confusion matrix analysis}

\begin{figure}[H]
    \centering
    \begin{subfigure}[b]{0.49\linewidth}
        \centering
        \includegraphics[width=\linewidth]{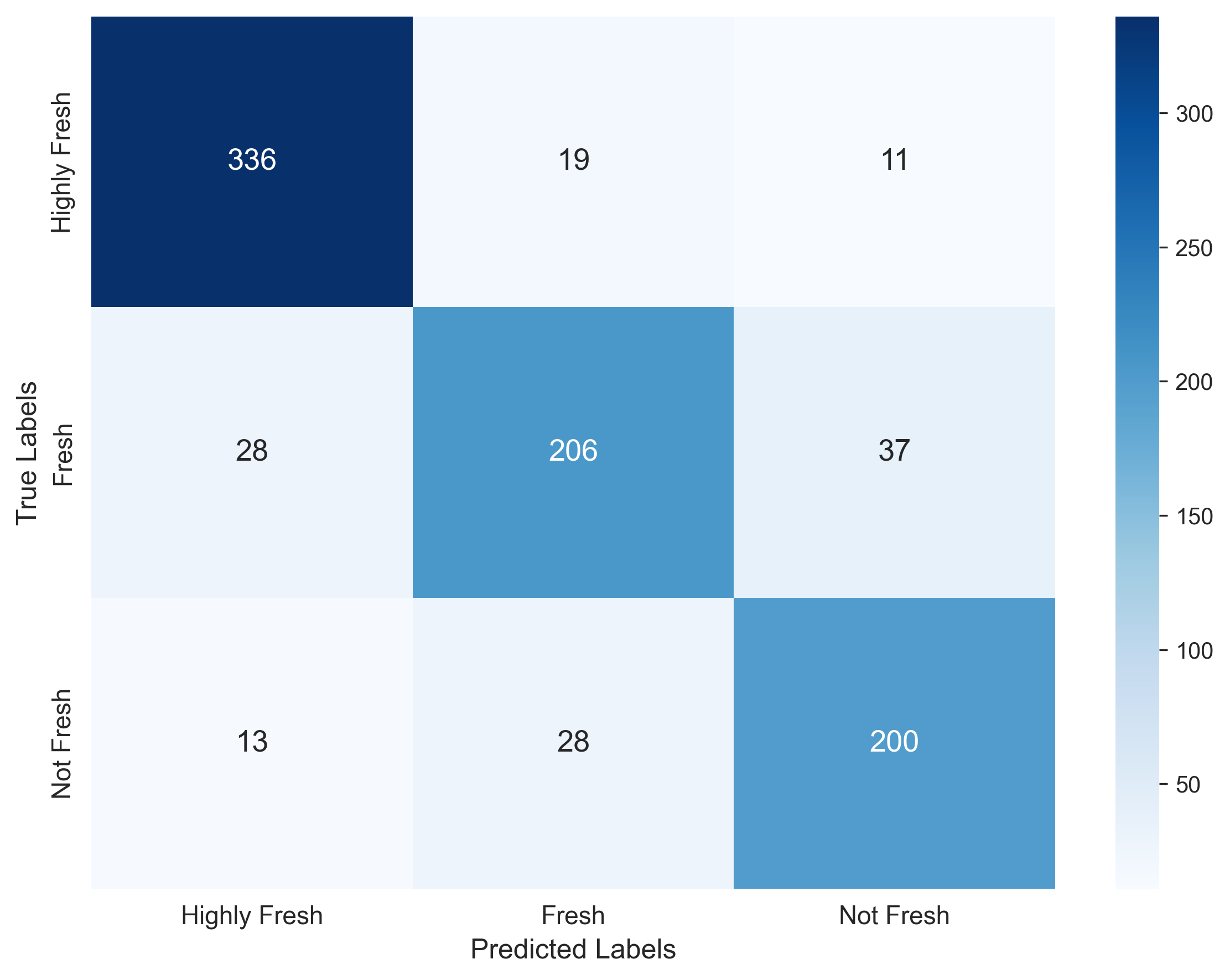}
        \caption{ConvNeXt-Base}
        \label{fig:cm_convnext}
    \end{subfigure}
    \hfill
    \begin{subfigure}[b]{0.49\linewidth}
        \centering
        \includegraphics[width=\linewidth]{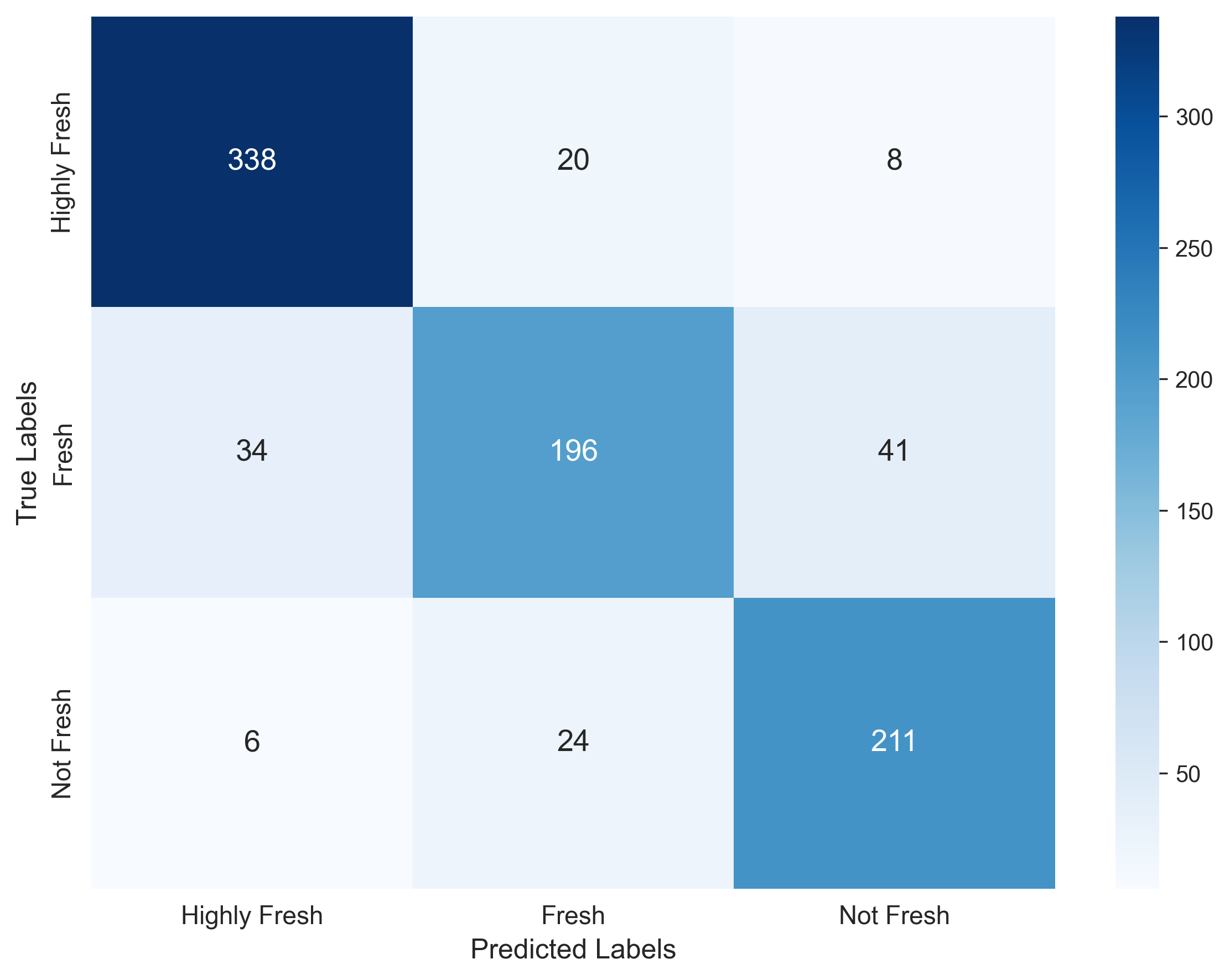}
        \caption{Swin-Tiny}
        \label{fig:cm_swin}
    \end{subfigure}
    \caption{Confusion matrices of ConvNeXt-Base and Swin-Tiny for fish freshness assessment.}
    \label{fig:cm_DL_comparison}
\end{figure}

To gain a clearer understanding of the specific inter-class misclassifications made by the models, we provide a detailed analysis of the Confusion Matrices (CM) for the two best-performing architectures: ConvNeXt-Base and Swin-Tiny, as shown in Figure \ref{fig:cm_DL_comparison}.

The analysis revealed that both models were highly effective at identifying the extreme classes. Swin-Tiny accurately classified 338 ``\textit{Highly fresh}'' samples, while ConvNeXt-Base performed comparably well by correctly identifying 336 samples. However, the models diverged in handling the critical boundary zones. ConvNeXt-Base was marginally better at identifying the difficult middle class, correctly classifying 206 ``\textit{Fresh}'' samples compared to Swin-Tiny's 196. At the same time, it demonstrated a weakness in risk assessment, as it misclassified 13 truly ``\textit{Not fresh}'' samples as ``\textit{Highly fresh}''. In contrast, Swin-Tiny was more robust against this critical failure, making the same error on only 6 samples. This suggests that while ConvNeXt-Base is slightly better at the subtle distinctions within the ``\textit{Fresh}'' category, Swin-Tiny provides superior reliability because it is far less likely to over-evaluate spoiled fish, which is a key requirement for safety-critical applications.

\subsubsection{Grad-CAM analysis}

\begin{figure}[H]
\centering
\includegraphics[width=1.0\textwidth]{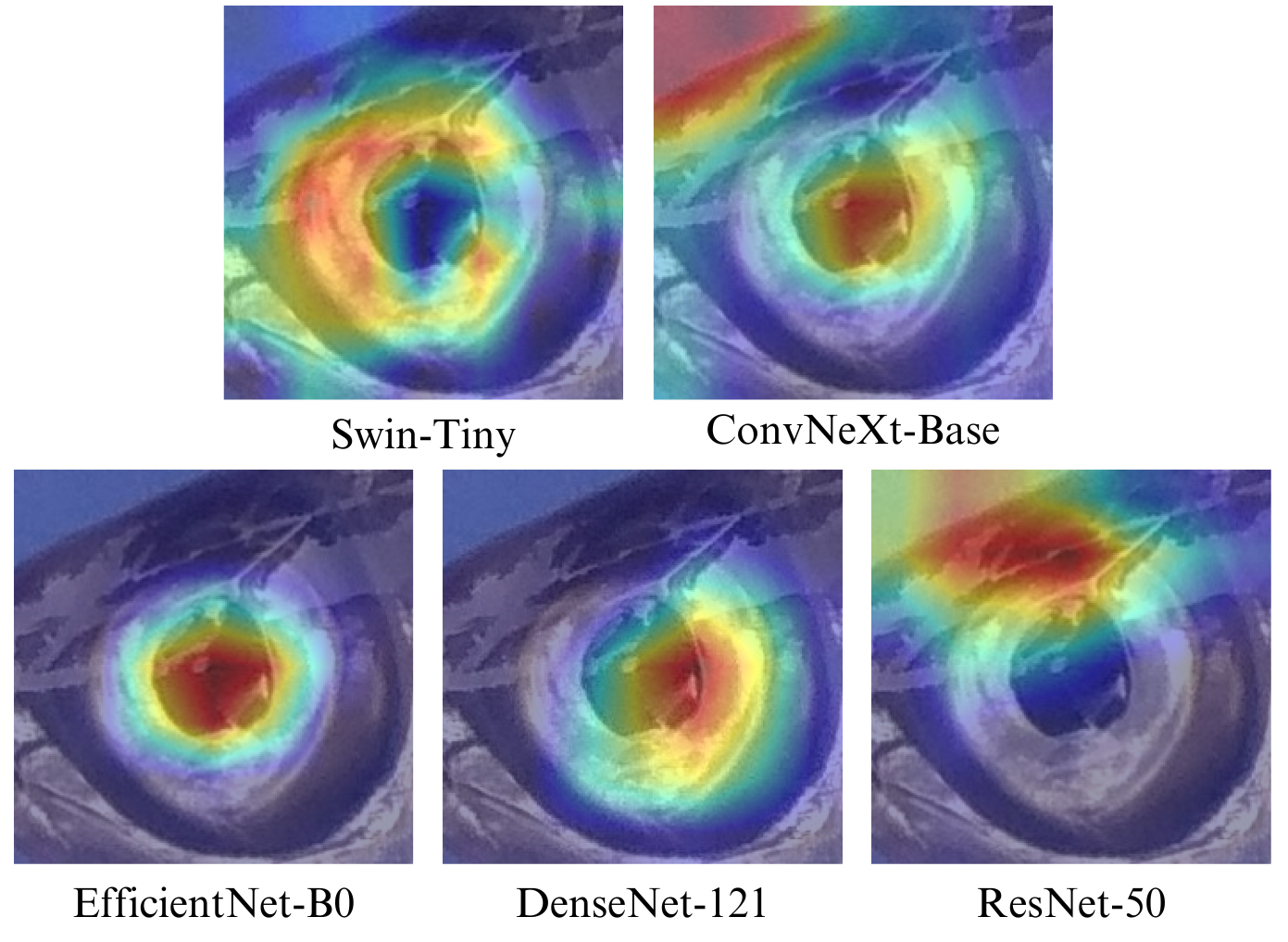} 
\caption{Grad-CAM visualizations highlighting the image regions attended to by different deep learning models during fish freshness classification.}
\label{fig:gradcam}
\end{figure}

To better understand how the deep learning models make decisions, we applied Gradient-weighted Class Activation Mapping (Grad-CAM). This technique highlights the specific regions in the images that most influence the models' predictions. The heatmaps shown in Figure \ref{fig:gradcam}  revealed important insights into both the internal reasoning of the models and the quality of the feature representations they generated. These visualizations show a clear relationship between how accurately a model focuses on relevant areas and its overall classification performance.

For the top-performing models, Swin-Tiny and ConvNeXt-Base, the Grad-CAM results are particularly informative. Swin-Tiny shows attention to both fine details of the central pupil and the larger structure of the eye. This ability to integrate local and global information allows the model to form a comprehensive understanding of fish freshness. ConvNeXt-Base also demonstrated focused attention, though slightly more diffuse, balancing detailed feature extraction with broader contextual understanding. This precise and balanced focus indicates that these models captured meaningful, discriminative features rather than random patterns, which contributes to their high accuracy.
In comparison, other models show clear limitations in their attention patterns. EfficientNet-B0 focused almost entirely on the pupil, which highlighted local textural changes but limited its awareness of surrounding structures, likely contributing to the high adjacency misclassifications. DenseNet-121 spread its attention more widely due to its dense connectivity. While this encouraged feature reuse, it appeared to dilute the focus on the most informative regions. ResNet-50 showed a scattered attention map with substantial background influence, suggesting difficulties in accurately localizing key areas.

These visualizations confirm the strength of transformer-based features, which we further exploit in hybrid setups below.

\subsection{Performance of full deep features}
\label{sec:performance_full_deep_features}

After evaluating the strong performance of deep learning architectures in Section~\ref{sec:performance_of_deep_learning_models}, we further investigated whether the discriminative representations learned by these backbones could be effectively exploited by classical machine learning algorithms, instead of relying on the original fully connected layers with a softmax output. By replacing the neural classification head with alternative ML classifiers, this stage aims to determine whether such algorithms can match or even surpass the discriminative capability of the learned deep models under identical experimental conditions.
Consequently, a hybrid learning strategy was adopted: feature representations were extracted from the middle-to-high layers of two major convolutional or transformer blocks of each fine-tuned backbone (e.g., ConvNeXt, Swin-Tiny, EfficientNet-B0, ResNet-50, and DenseNet-121), thereby capturing complementary levels of abstraction. These feature maps were then processed using Global Average Pooling (GAP) to obtain compact, fixed-length embeddings, which subsequently served as inputs to a suite of traditional machine learning classifiers, including LR, KNN, SVM, ANN, RF, ET, LGBM, and CatBoost. All experiments in this section were conducted under the same holdout protocol (64\% for training, 16\% for validation, and 20\% for testing). This unified setup ensures a fair and consistent comparison between the DL models and the proposed hybrid feature-based approach.

Based on the previous analysis, the Swin-Tiny model, which achieved the best overall performance among the evaluated architectures, was selected as the primary backbone for feature extraction. 
To further investigate the effect of feature abstraction, both middle-level and high-level embeddings were evaluated using identical machine learning classifiers. As shown in Table~\ref{tab:ml_full_features}, features extracted from higher layers consistently yielded better results across all classifiers. While the best performance at the middle level was achieved by SVM with an accuracy of 84.97\%, all models benefited from the higher-level representations, where ET achieved the highest accuracy of 85.88\%, marginally outperforming the Swin-Tiny model (84.85\%). This observation suggests that deeper representations are more discriminative, capturing richer semantic information that is crucial for distinguishing subtle variations in fish freshness. In contrast, middle-level features--although containing finer textural details--may lack the semantic consistency required for optimal classification.

\begin{table}[H]
\caption{Performance of ML classifiers on deep features extracted from Swin-Tiny \\(Stage 3 vs. Stage 4) (\%).}
\label{tab:ml_full_features}
\centering
\footnotesize
\begin{tabular}{|l|cccc|cccc|}
\hline
\multirow{2}{*}{\textbf{Model}} & \multicolumn{4}{c|}{\textbf{Stage 3 (middle-level features)}} & \multicolumn{4}{c|}{\textbf{Stage 4 (high-level features)}} \\
\cline{2-9}
 & \textbf{ACC} & \textbf{Recall} & \textbf{Precision} & \textbf{F1} & \textbf{ACC} & \textbf{Recall} & \textbf{Precision} & \textbf{F1} \\
\hline
LR       & 81.89 & 81.89 & 81.87 & 81.82 & 85.19 & 85.19 & 85.34 & 85.24 \\
KNN      & 83.03 & 81.03 & 83.12 & 83.10 & 85.65 & 85.65 & 85.67 & 85.64 \\
SVM      & \textbf{84.97} & 84.97 & 84.99 & \textbf{84.98} & 85.65 & 85.65 & 85.58 & 85.59 \\
ANN      & 82.23 & 82.23 & 82.36 & 82.26 & 85.54 & 85.54 & 85.67 & 85.57 \\
RF       & 79.95 & 79.95 & 79.75 & 79.69 & 85.31 & 85.31 & 85.48 & 85.38 \\
ET       & 80.75 & 80.75 & 80.52 & 80.43 & \textbf{85.88} & \textbf{85.88} & \textbf{85.91} & \textbf{85.89} \\
LGBM & 81.66 & 81.66 & 81.55 & 81.59 & 85.08 & 85.08 & 85.15 & 85.10 \\
CB & 80.64 & 80.64 & 80.50 & 80.54 & 84.28 & 84.28 & 84.35 & 84.30 \\
\hline
\end{tabular}
\end{table}

\begin{figure}
    \centering
    \includegraphics[width=0.8\linewidth]{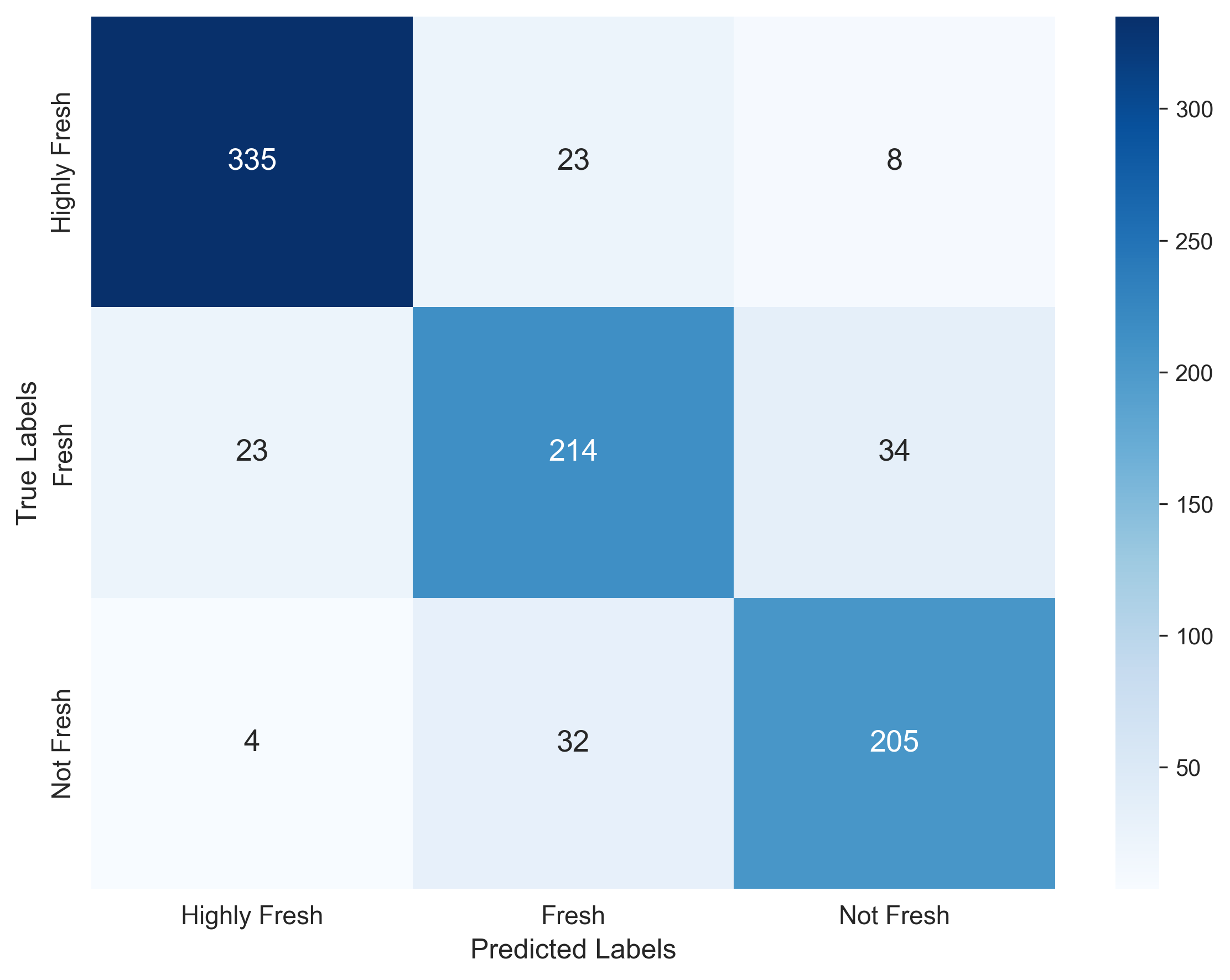}
   \caption{Confusion matrix of ET using features extracted from Stage 4 of Swin-Tiny}
    \label{fig:CM_Swin_ET_full}
\end{figure}

In addition, we analyzed the confusion matrix of the best-performing configuration (Stage 4 features + ET). As shown in the confusion matrix (Figure \ref{fig:CM_Swin_ET_full}), this combination achieved balanced and consistent performance across all three freshness categories. The model correctly identified most samples in each class, with accuracy rates of 91.5\% for Highly fresh, 78.9\% for Fresh, and 85.1\% for Not fresh, resulting in an overall accuracy of 85.88\%. Compared to the Swin-Tiny model, this approach improved the classification of the Fresh class by approximately 6.6\%, indicating that the tree-based classifier can better handle subtle variations between adjacent categories. Although a slight decrease was observed for the Not Fresh class, the results remain competitive and more evenly distributed. These findings suggest that the Stage 4 embeddings of Swin-Tiny provide sufficiently discriminative representations, and when combined with ET, they enhance generalization performance.

Following the selection of the best-performing backbone, we also conducted experiments using feature embeddings extracted from all other deep architectures to verify the generality of this hybrid approach. The same feature-extraction and training protocol was consistently applied across ConvNeXt, EfficientNet, ResNet-50, and DenseNet to ensure comparability.
Importantly, the improvement observed with Swin-Tiny was not an isolated case. As shown in Table~\ref{tab:comparative_dl_features}, the hybrid deep feature–based learning strategy consistently enhanced performance across all evaluated architectures. For instance, ConvNeXt-Base improved from 84.51\% to 85.19\%, while more traditional CNNs showed even greater gains--ResNet-50 increased from 80.07\% to 83.14\%, and DenseNet-121 from 80.75\% to 83.49\%. These results confirm that the fully connected layers in DL models may not always provide the most effective decision boundaries, and that classical machine learning classifiers can more efficiently exploit the rich representations generated by deep backbones.

\begin{table}[H]
\caption{Maximum accuracy achieved by each deep learning model using features from its optimal architectural level (\%).}
\label{tab:comparative_dl_features}
\footnotesize
\hspace*{-1.0cm}
\centering
\begin{tabular}{|l|c|c|c|c|c|}
\hline
\textbf{Deep feature} & \textbf{Best classifier} & \textbf{Feature source} & \textbf{Dimension} & \textbf{Best Acc} & \textbf{Impact} \\
\hline
ConvNeXt-Base   & ET       &      Stage 4 (Final stage)    & 1024  & 85.19 & +0.68\% \\
\hline
Swin-Tiny       & ET       &     Stage 4 (Final stage)    & 768   & \textbf{85.88} & +1.03\% \\
\hline
EfficientNet-B0 & SVM      & Block 7         & 320   & 82.69 & +1.37\% \\
\hline
DenseNet-121    & KNN      & DenseBlock 4    & 1024  & 83.49 & +2.74\% \\
\hline
ResNet-50       & LGBM     & Layer 4         & 2048  & 83.14 & +3.07\% \\
\hline
\end{tabular}
\end{table}

A crucial observation from Table~\ref{tab:comparative_dl_features} is that, across all evaluated architectures, the most effective features were consistently extracted from the higher-level representation layers. This pattern reinforces the earlier finding from Swin-Tiny that deeper embeddings provide superior discriminative power compared to those from intermediate levels. 
ET achieved the best results with transformer-derived representations extracted from Swin-Tiny (768 features) and ConvNeXt-Base (1024 features), highlighting its capacity to model complex, high-level dependencies. KNN performed effectively with DenseNet-121 features (1024 features), consistent with DenseNet's feature-reuse mechanism that forms compact and locally coherent clusters in the embedding space. LGBM showed strong performance with ResNet-50 features (2048 features), demonstrating its robustness in capturing non-linear relationships within high-dimensional data. Finally, SVM achieved the best results with EfficientNet-B0 features (320 features), as its compound scaling and MBConv design produce compact and regularized feature spaces that enhance margin-based separation.

\subsection{Performance of selected deep features}
\label{sec:performance_of_selected_deep_features}

The analysis in Section \ref{sec:performance_full_deep_features} demonstrated that high-level features extracted from deep architectures, when used with classical machine learning classifiers, are highly discriminative. However, their high dimensionality (320–2048 features) can introduce noise, redundancy, and low-predictive-power features, increasing computational cost and potentially limiting performance. This section investigates whether a smaller, more informative subset of features can improve overall performance. Three embedded feature selection methods-LGBM (boosting), RF (bagging), and Lasso regression (L1)-were applied to filter irrelevant predictors. The impact of dimensionality reduction was examined stepwise by progressively retaining the top 50\%, 40\%, 30\%, 20\%, 10\%, and 5\% of features. This gradual reduction enabled us to observe performance trends and identify the optimal trade-off between feature subset size and predictive accuracy before a noticeable drop occurred. The analysis focuses on features extracted from the Swin-Tiny backbone, which achieved the best performance in previous experiments, to evaluate the impact of feature selection.

\subsubsection{Impact of feature selection using LGBM (boosting)}

The feature selection process with LGBM, a boosting-based method, produced particularly informative results, as shown in Table \ref{tab:feature_selection_impact_swin_stage4_using_LGBM}. Reducing the feature dimension did not degrade performance; the highest accuracy of 85.99\% was achieved in three cases: RF with 77 features, ET with 231 features, and ET with 384 features. Among these, RF with only 77 features represents the most compact subset while retaining maximum discriminative power, yielding a +0.11\% improvement over the baseline accuracy of 85.88\%. This corresponds to a 90\% reduction in features, highlighting the efficiency of a small, informative subset. These results demonstrate that LGBM effectively filters out redundant or noisy features, leading to a more robust and computationally efficient model.

\begin{table}[H]
\footnotesize
\caption{Impact of feature selection using LGBM (boosting) (\%).}
\label{tab:feature_selection_impact_swin_stage4_using_LGBM}
\centering
\begin{tabular}{|l|c|c|c|c|}
\hline
\textbf{Feature subset} & \textbf{Dimension} & \textbf{Best classifier} & \textbf{Accuracy} & \textbf{Impact} \\
\hline
100\% (Full set)                 & 768 & ET & 85.88 & Baseline \\
\hline
50\%                             & 384 & ET & \textbf{85.99} & +0.11\% \\
\hline
40\%                             & 308 & KNN    & 85.65 & -0.23\% \\
\hline
30\%                             & 231 & ET & \textbf{85.99} & +0.11\% \\
\hline
20\%                             & 154 & SVM    & 85.88 & +0.00\% \\
\hline
10\%                             & 77  & RF     & \textbf{85.99} & +0.11\% \\
\hline
5\%                              & 39  & RF     & 85.54 & -0.34\% \\
\hline
\end{tabular}
\end{table}

\subsubsection{Impact of feature selection using RF (bagging)}

Feature selection using RF (bagging) did not provide any performance gains, as shown in Table \ref{tab:feature_selection_impact_swin_stage4_using_RF}. While reducing the number of features did not substantially harm classification accuracy, no subset outperformed the full feature set. The peak accuracy of 85.88\% was achieved with the top 30\% of features (231 dimensions), matching the baseline. Other subsets led to minor decreases in accuracy, ranging from -0.23\% to -0.46\%. These findings indicate that, despite the strong generalization capability of RF, its feature importance estimation is less capable than LGBM's in identifying a concise set of discriminative features for the FFE dataset.

\begin{table}[H]
\footnotesize
\caption{Impact of feature selection using RF (bagging) (\%).}
\label{tab:feature_selection_impact_swin_stage4_using_RF}
\centering
\begin{tabular}{|l|c|c|c|c|}
\hline
\textbf{Feature subset} & \textbf{Dimension} & \textbf{Best classifier} & \textbf{Accuracy}  &    \textbf{Impact}        \\
\hline
100\% (Full set)        &        768         &             ET           &        85.88       &          Baseline         \\
\hline
50\%                    &        384         &            SVM           &        85.42       &          -0.46\%          \\
\hline
40\%                    &        308         &            KNN           &        85.65       &          -0.23\%          \\
\hline
30\%                    &        231         &            KNN           &        85.88       &            0\%            \\
\hline
20\%                    &        154         &            RF            &        85.42       &          -0.46\%          \\
\hline
10\%                    &        77          &            SVM           &        85.54       &          -0.34\%          \\
\hline
5\%                     &        39          &            KNN           &        85.42       &          -0.46            \\
\hline
\end{tabular}
\end{table}

\subsubsection{Impact of feature selection using Lasso regression (L1)}

Lasso regression (L1), a regularization-based method, was less suitable for this task, as shown in Table \ref{tab:feature_selection_impact_swin_stage4_using_L1}. Unlike the tree-based methods, L1 consistently reduced classification accuracy. Performance generally declined as the number of retained features decreased. Even the largest subset, the top 50\%, resulted in a -0.34\% drop. Notably, with the top 10\% of features (77 dimensions), the decrease was smaller (-0.12\%), suggesting that Lasso can preserve some discriminative information in very compact subsets. Overall, these results indicate that the relationships between deep features and fish freshness classes are highly complex and non-linear, and Lasso's linear penalization likely removed features crucial for capturing these patterns, which are better handled by non-linear classifiers such as ET and SVM.

\begin{table}[H]
\footnotesize
\caption{Impact of feature selection using Lasso Regression (L1) (\%).}
\label{tab:feature_selection_impact_swin_stage4_using_L1}
\centering
\begin{tabular}{|l|c|c|c|c|}
\hline
\textbf{Feature subset} & \textbf{Dimension} & \textbf{Best classifier} & \textbf{Accuracy}  &    \textbf{Impact}   \\
\hline
100\% (Full set)        &        768         &           RT         &        85.88       &          Baseline         \\
\hline
50\%                    &        384         &            SVM           &        85.54       &           -0.34\%         \\
\hline
40\%                    &        308         &            SVM           &        85.54       &           -0.34\%         \\
\hline
30\%                    &        231         &            KNN           &        85.42       &           -0.46\%         \\
\hline
20\%                    &        154         &           ET         &        85.31       &           -0.57\%         \\
\hline
10\%                    &        77          &            SVM           &        85.76       &           -0.12\%          \\
\hline
5\%                     &        39          &            KNN           &        84.97       &           -0.94\%          \\
\hline
\end{tabular}
\end{table}

We can observe that LGBM consistently yields the most effective feature subsets, which can be attributed to its distinctive algorithmic properties. LGBM, as a gradient boosting-based method, provides more informative feature selection compared to RF and L1-based methods due to several key factors. First, unlike RF which evaluates feature importance primarily based on individual tree splits and can be biased toward features with more levels or higher cardinality, LGBM considers the cumulative contribution of each feature across all boosting iterations, capturing subtle interactions among features. Second, LGBM's leaf-wise growth strategy with gradient-based optimization prioritizes splits that maximize loss reduction, enabling it to identify features that are most discriminative for the target task, even if their individual effects are small. In contrast, L1 regularization performs feature selection by shrinking coefficients to zero in a linear model, which may overlook nonlinear relationships or interactions among features. As a result, LGBM selects a subset that is both compact and highly informative, effectively filtering out redundant or noisy features while retaining maximum predictive power, as evidenced by the improved accuracy with a substantially reduced feature set.

\subsubsection{Observed effects of LGBM feature selection across deep architectures}

To provide a broader perspective, Table~\ref{tab:best_feature_subsets} shows the best-performing feature subsets obtained through LGBM-based selection across all evaluated deep architectures. A consistent trend emerges: for most models--Swin-Tiny, ConvNeXt-Base, DenseNet-121, and ResNet-50--the top 10\% of selected features achieved the highest accuracy, outperforming the full feature sets. For instance, ConvNeXt-Base attained its best result (85.31\%) with an SVM classifier using 103 features, indicating that LGBM effectively preserves nonlinear discriminative information. DenseNet-121 performed best with ET (83.60\%), while ResNet-50 reached 83.49\% with LGBM, reflecting robustness in moderately high-dimensional spaces. The only exception was EfficientNet-B0, whose accuracy slightly decreased (82.67\% $\rightarrow$ 82.35\%) due to its already compact 192-dimensional representation, where further reduction removed informative features.

\begin{table}[H]
\footnotesize
\caption{Best-performing feature subsets selected by LGBM (boosting) for each deep learning model (\%)}
\label{tab:best_feature_subsets}
\centering
\begin{tabular}{|l|c|c|c|c|}
\hline
\textbf{Model}  & \textbf{Feature subset} & \textbf{Dimension} & \textbf{Best classifier} & \textbf{Accuracy} \\
\hline
Swin-Tiny       &               10\%               &             77             &            RF            &      \textbf{85.99}    \\
\hline
ConvNeXt-Base   &               10\%               &            103             &           SVM            &          85.31         \\
\hline
DenseNet-121    &               10\%               &            103             &            ET            &          83.60          \\
\hline
ResNet-50       &               10\%               &            205             &           LGBM           &          83.49          \\
\hline
EfficientNet-B0 &               50\%               &            160             &           KNN            &          82.35          \\
\hline
\end{tabular}
\end{table}

\subsection{Analysis of model parameters and training time}

\begin{figure}[H]
\centering
\includegraphics[width=1.0\textwidth]{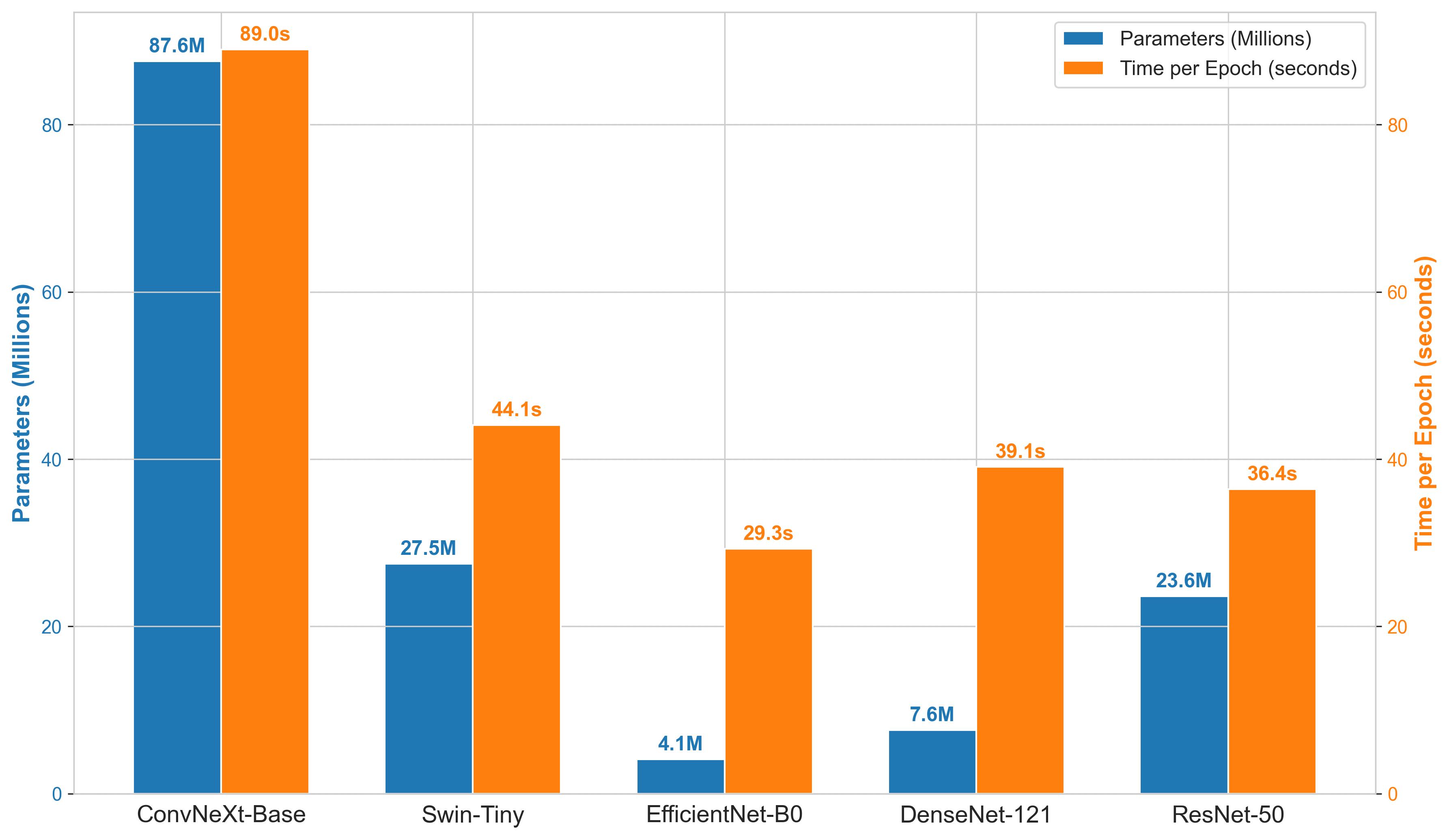}
\caption{Comparison of model complexity and training time per epoch across vision architectures.}
\label{fig:compare_param_training_time}
\end{figure}

In addition to predictive accuracy, this study conducted an analysis of model complexity and computational cost, as illustrated in Figure~\ref{fig:compare_param_training_time}, to emphasize the trade-offs between different architectures. The findings indicate that parameter count is not the sole factor influencing computational expense; rather, architectural design plays a critical role in overall efficiency. To quantify this, we computed parameter efficiency as parameters (in millions) divided by training time per epoch (in seconds), which yielded units of M/s to measure processing speed.

The comparison highlights the contrast between complexity and efficiency. ConvNeXt-Base (87.6M parameters, 89.0s per epoch, efficiency $\approx$ 0.98 M/s) exemplifies a highly complex design that achieves strong accuracy but demands considerable resources, making it resource-intensive and potentially limiting its practicality for deployment on constrained hardware. In contrast, EfficientNet-B0 (4.1M parameters, 29.3s per epoch, efficiency $\approx$ 0.14 M/s) represents the opposite extreme, offering remarkable efficiency at the cost of reduced accuracy.

The classic architectures provide additional insight into these dynamics. DenseNet-121 (7.6M parameters, 39.1s per epoch, efficiency $\approx$ 0.19 M/s) has far fewer parameters than ResNet-50 (23.6M parameters, 36.4s per epoch, efficiency $\approx$ 0.65 M/s), yet it requires longer training time due to its feature concatenation operations. This outcome underscores that architectural choices, such as DenseNet's dense connectivity, can outweigh raw parameter efficiency in terms of computational overhead.

Within this context, Swin-Tiny emerged as the most balanced solution. With 27.5M parameters and a training time of 44.1s per epoch (efficiency $\approx$ 0.62 M/s), it represents a moderate resource investment while achieving the highest accuracy among all models. It represented fewer computational resources than ConvNeXt-Base (about 31\% fewer parameters and 50\% less time per epoch), while still outperforming EfficientNet-B0 and DenseNet-121 in both accuracy and efficiency (e.g., 4.4 times higher than EfficientNet-B0's 0.14 M/s), with only a modest increase in cost compared to ResNet-50. This combination of leading accuracy and manageable resource requirements identifies Swin-Tiny as the most compelling backbone for developing models that are both high-performing and practically deployable.

Based on these observations, the analysis highlighted that model selection is a multi-dimensional decision involving more than predictive accuracy. While complex models like ConvNeXt-Base can achieve high accuracy and lightweight models like EfficientNet-B0 offer efficiency, neither extreme is optimal on its own. A balanced architecture such as Swin-Tiny provided a practical compromise, delivering state-of-the-art accuracy with manageable computational cost--particularly beneficial for applications in resource-limited environments like edge computing devices. This makes it suitable for further hybrid model development in real-world scenarios.

\subsection{Comparison with existing studies using the same dataset}

\begin{figure}[H]
\centering
\includegraphics[width=1\textwidth]{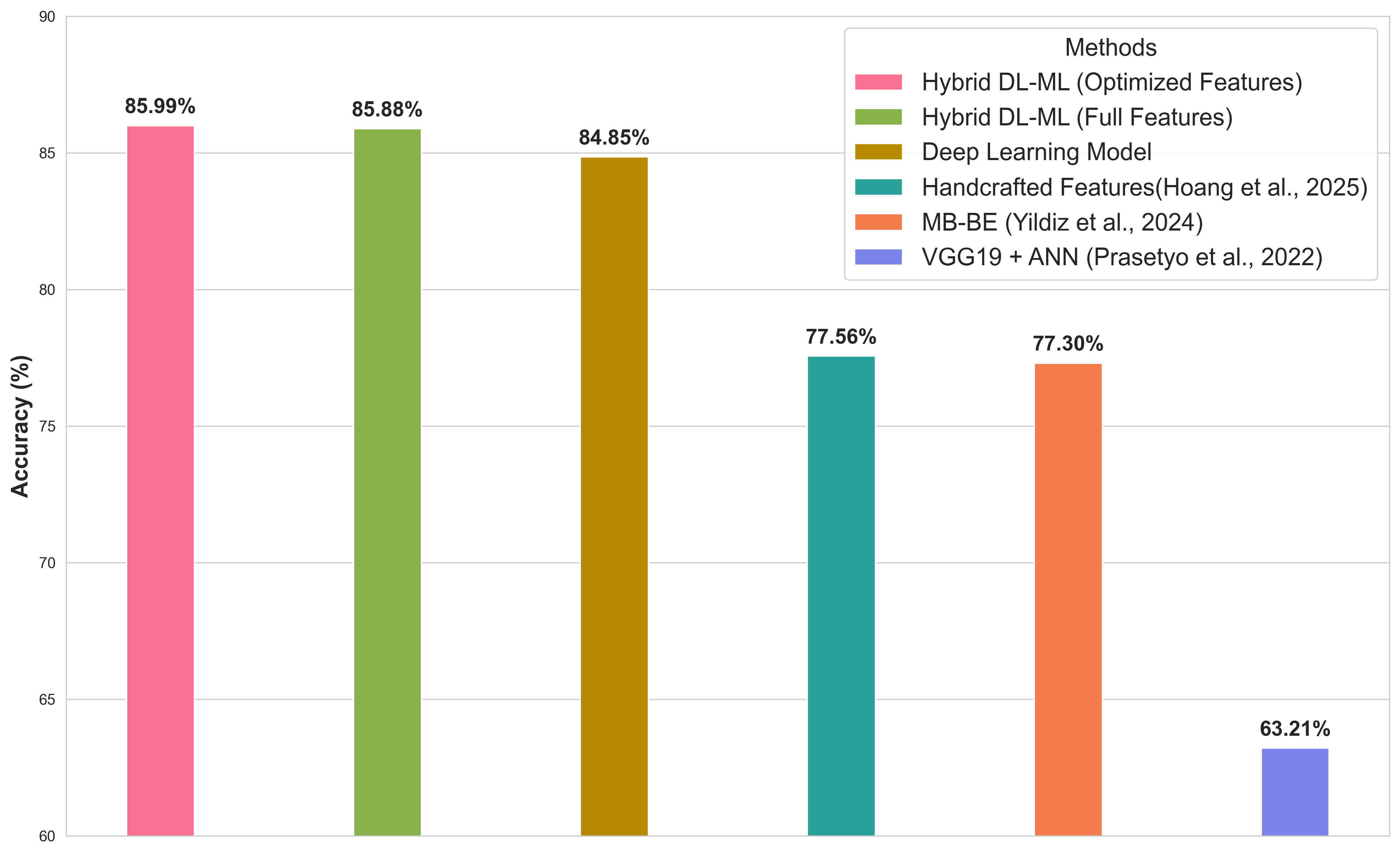} 
\caption{Accuracy comparison of the proposed strategies with previous studies on the FFE dataset.}
\label{fig:comparation_methods}
\end{figure}

To emphasize the contributions of this study, a comparative analysis was conducted against previously published works using the same \textit{Freshness of the Fish Eyes (FFE)} dataset. As illustrated in Figure~\ref{fig:comparation_methods}, the proposed methods demonstrate clear improvements in both classification accuracy and robustness compared with existing approaches.

The best-performing model, the proposed hybrid method (features from Swin-Tiny + ET), achieved 85.99\% accuracy, surpassing the results of~\cite{Hoang2025} by 8.43\%,~\cite{Yildiz2024} by 8.69\%, and~\cite{Prasetyo2022} by 22.78\%. Even the baseline deep learning model (Swin-Tiny) attained 84.85\% accuracy, exceeding all previous results. These outcomes highlight the advantage of using a modern vision transformer backbone and the benefit of combining complementary feature representations.

The superiority of the proposed framework stems from its ability to address the main limitations of earlier studies. The approach by~\cite{Hoang2025} is limited by its reliance on handcrafted features, which requires significant domain expertise for engineering and may fail to capture the abstract visual cues that deep learning models learn automatically. Similarly, the hybrid model proposed by~\cite{Yildiz2024} was constrained by its use of conventional convolutional backbones (VGG19 and SqueezeNet), limiting its capacity to model long-range dependencies. 
In contrast, the lightweight MB-BE architecture of ~\cite{Prasetyo2022} emphasized computational efficiency at the expense of representational capacity. Its compact design likely constrained the ability to learn complex, fine-grained visual cues, particularly given the limited size and variability of the training dataset.

The proposed framework achieves higher predictive accuracy while providing a balanced and generalizable representation of visual information. It captures both semantic patterns and physical cues, offering a reliable foundation for future research and practical applications in automated fish freshness assessment.

\subsection{Discussion}

The proposed hybrid framework substantially improves the accuracy of fish freshness assessment on the FFE dataset, achieving 85.99\% using the Swin-Tiny backbone combined with an ET classifier and LGBM-based feature selection. This notable improvement highlights the effectiveness of integrating deep visual representations with classical ML and embedded feature optimization. The three-stage design plays a crucial role in this success: fine-tuning captures rich semantic features; traditional classifiers define sharper and more flexible decision boundaries than a softmax layer; and feature selection isolates compact, informative subsets that reduce noise and enhance generalization.

\begin{figure}[H]
\centering
\includegraphics[width=1.0\textwidth]{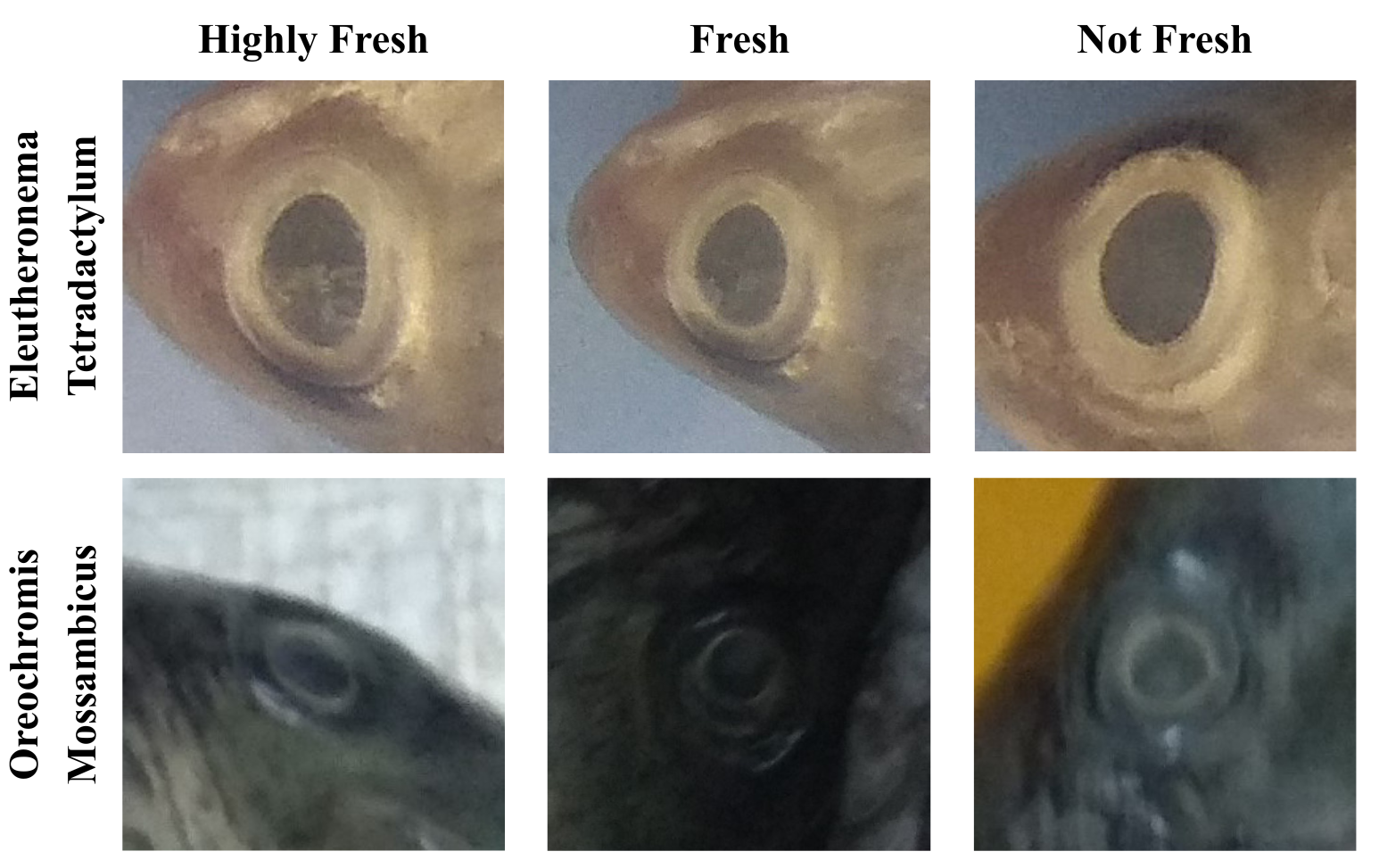} 
\caption{Examples of ambiguous or misclassified fish-eye images from the FFE dataset (\textit{Highly fresh}, \textit{Fresh}, and \textit{Not fresh}) for \textit{Eleutheronema tetradactylum} and \textit{Oreochromis mossambicus}.}
\label{fig:samples_for_discussion}
\end{figure}

Despite the promising results, the proposed framework is constrained by the inherent limitations of the FFE dataset, which comprises only 4,390 images. This relatively small scale and idealized setup may limit the model's robustness to real-world variations, such as diverse lighting conditions, species diversity, or environmental noise encountered in industrial seafood processing environments. Figure~\ref{fig:samples_for_discussion} illustrates representative ambiguous samples in which the visual differences between freshness classes are extremely subtle, making it challenging to distinguish class boundaries even for human observers.

In addition to dataset-related constraints, several methodological limitations should be noted. The three-stage, decoupled design increases implementation complexity, as each step--fine-tuning, feature extraction, feature selection, and classifier training--requires separate setup and optimization. While this modularity enhances interpretability and debugging flexibility, it also demands greater technical and computational resources, potentially hindering deployment in real-time or resource-constrained environments.

Another limitation involves the exclusive use of deep features extracted from the Global Average Pooling (GAP), without incorporating handcrafted features. Deep features effectively capture high-level semantics but lack interpretable physical attributes. For instance, handcrafted color features (e.g., HSV histograms) could characterize eye redness, while texture descriptors (e.g., gray-level co-occurrence matrices) could quantify lens opacity. Combining both through feature fusion could yield a more comprehensive representation, integrating abstract contextual information with fine-grained physical cues.

Collectively, these findings establish a solid methodological foundation for automated fish freshness assessment and indicate promising directions for developing models that are both accurate and interpretable for real-world implementation.

\section{Conclusion}
\label{sec:conclusion}

This study presented a comprehensive framework for optimizing deep feature representations in fish freshness assessment. The framework systematically integrates deep backbone fine-tuning, multi-level feature extraction, classical machine learning classification, and embedded feature selection. Several state-of-the-art architectures, including ResNet-50, DenseNet-121, EfficientNet-B0, ConvNeXt-Base, and Swin-Tiny, were fine-tuned and evaluated to establish a robust experimental baseline. Different ML classifiers were then applied to the extracted embeddings, while three embedded selection methods - LGBM, RF, and L1 - were compared to identify the most compact and discriminative feature subsets.
Experimental results showed that LGBM-based feature selection consistently enhanced accuracy across architectures, with the best configuration (Swin-Tiny features and ET classifier) achieving 85.99\% accuracy on the FFE dataset. This performance surpasses existing approaches, such as~\cite{Hoang2025},~\cite{Yildiz2024} and~\cite{Prasetyo2022}, by 8.43\% and 22.78\%, respectively. These results confirm that combining modern vision transformers with traditional ensemble classifiers and boosting-based feature selection yields a powerful and generalizable strategy for visual quality evaluation tasks.
Future research will focus on developing hybrid representations that combine deep and handcrafted features to capture both semantic and physical attributes of fish freshness. Further validation on larger and more diverse real-world datasets is expected to improve robustness and practical applicability.

\section*{Funding:} This research received no funding.

\section*{Data availability:} 

The datasets generated and analyzed during the current study are available in the Mendeley data repository~\citep{MendeleyFFEData}.

\section*{Author contribution:}

    \textbf{Phi-Hung Hoang}: Writing - review \& editing, Writing - original draft, Data curation, Visualization, Validation, Software, Methodology, Investigation. \textbf{Nam-Thuan Trinh}: Writing - review \& editing, Writing - original draft, Data curation, Visualization, Validation, Software, Investigation. \textbf{Van-Manh Tran}: Writing - review \& editing, Writing - original draft, Data curation, Software, Investigation. \textbf{Thi-Thu-Hong Phan}: Conceptualization, Methodology, Supervision, Validation, Writing - original draft, Writing - review \& editing.

\section*{Declaration of interests:}
 
The authors declare that they have no known competing financial interests or personal relationships that could have appeared to influence the work reported in this paper.

\section*{Declaration of generative AI and AI-Assisted technologies in the writing process:}

During the preparation of this manuscript, Grammarly and generative large language models were employed to improve grammar and refine wording for clarity. The authors subsequently reviewed and revised the content as necessary and take full responsibility for the final version of the manuscript.

\end{document}